\pdfoutput=1

\documentclass[11pt]{article}

\usepackage{acl}

\usepackage{times}
\usepackage{latexsym}

\usepackage[T1]{fontenc}

\usepackage[utf8]{inputenc}

\usepackage{microtype}
\usepackage{amsmath}
\usepackage{graphicx}
\usepackage{booktabs}
\usepackage{multirow}
\usepackage{amssymb}
\usepackage{paralist}

\title{Controllable User Dialogue Act Augmentation for Dialogue State Tracking}

\author{
Chun-Mao Lai\thanks{\hspace*{0.5em}Equal contribution.}\quad
{Ming-Hao Hsu}\footnotemark[1]\quad
Chao-Wei Huang\quad
Yun-Nung Chen \\ 
National Taiwan University, Taipei, Taiwan \\
\texttt{\{b09901186, b09502138\}@ntu.edu.tw} \\
\texttt{f07922069@csie.ntu.edu.tw} \hspace{1.5em}  
\texttt{y.v.chen@ieee.org}
}

\begin{document}
\maketitle
\begin{abstract}
Prior work has demonstrated that data augmentation is useful for improving dialogue state tracking.
However, there are many types of user utterances, while the prior method only considered the simplest one for augmentation, raising the concern about poor generalization capability.
In order to better cover diverse dialogue acts and control the generation quality, this paper proposes controllable user dialogue act augmentation (CUDA-DST) to augment user utterances with diverse behaviors. 
With the augmented data, different state trackers gain improvement and show better robustness, achieving the state-of-the-art performance on MultiWOZ 2.1.\footnote{The source code is available at \url{https://github.com/MiuLab/CUDA-DST}.}

\end{abstract}

\section{Introduction}
Dialogue state tracking (DST) serves as a backbone of task-oriented dialogue systems~\cite{chen2017deep}, where it aims at keeping track of user intents and associated information in a conversation.
The dialogue states encapsulate the required information for the subsequent dialogue components.
Hence, an accurate DST module is crucial for a dialogue system to perform successful conversations.


Recently, we have seen tremendous improvement on DST, mainly due to the curation of large datasets~\cite{budzianowski2018multiwoz,eric-etal-2020-multiwoz,rastogi2020towards} and many advanced models. 
They can be broadly categorized into 3 types: span prediction, question answering, and generation-based models.
The question answering models define natural language questions for each slot to query the model for the corresponding values~\cite{gao-etal-2020-machine,li-etal-2021-zero}.
\citet{wu-etal-2019-transferable} proposed TRADE to perform zero-shot transfer between multiple domains via slot-value embeddings and a state generator.
SimpleTOD~\cite{hosseini2020simple} combines all components in a task-oriented dialogue system with a pre-trained language model.
Recently, TripPy~\cite{heck-etal-2020-trippy} 
categorizes value prediction into 7 types, and designs different prediction strategies for them.
This paper focuses on generalized augmentation covering all categories.

Another research line leverages data augmentation techniques to improve performance~\cite{song2021data,yin2020dialog,summerville-etal-2020-tame,kim-etal-2021-neuralwoz}.
Most prior work used simple augmentation techniques such as word insertion and state value substitution.
With recent advances in pre-trained language models~\cite{devlin-etal-2019-bert,radford2019language,JMLR:v21:20-074}, generation-based augmentation has been proposed~\cite{kim-etal-2021-neuralwoz,li2020coco}.
These methods have demonstrated impressive improvement and zero-shot adaptability~\cite{yoo-etal-2020-variational,campagna2020zero}, while our work focuses on data augmentation with in-domain data.

\begin{figure*}
    \centering
    \includegraphics[width=\textwidth]{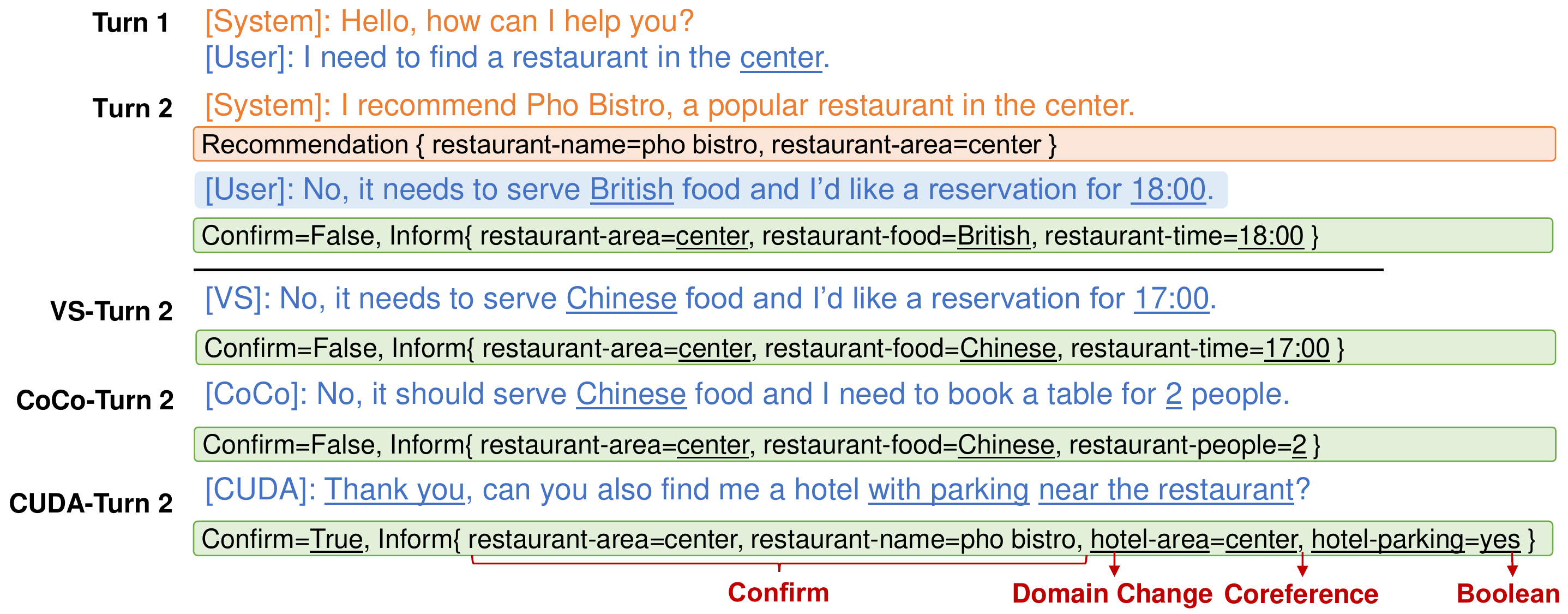}
    \caption{Augmented user utterances with the associated user dialogue acts and states from three methods.}
    \label{fig:example}
\end{figure*}

The closest work is CoCo~\cite{li2020coco}, a framework that generates user utterances given augmented dialogue states.
The examples are shown in Figure~\ref{fig:example}, where the main differences between CoCo and ours are that
1) CoCo only augments user utterances in slot and value levels, but dialogue acts and domains are fixed, making augmented data limited. Our method can augment reasonable user utterances with diverse dialogue acts and domain switching scenarios.
2) Boolean slots and referred slots are not handled by CoCo due to its higher complexity, while our approach can handle all types of values for better generalization.

This paper proposes \textbf{CUDA-DST} (\textbf{C}ontrollable \textbf{U}ser \textbf{D}ialogue \textbf{A}ct augmentation), 
a generalized framework of generation-based augmentation for improving DST.
Our contribution is 2-fold:
\begin{compactitem}
    \item We present CUDA which generates diverse user utterances via controllable user dialogue acts augmentation.
    \item Our augmented data helps most DST models improve their performance. Specifically, CUDA-augmented TripPy model achieves the state-of-the-art result on MultiWOZ 2.1.
\end{compactitem}


\section{Controllable User Dialogue Act Augmentation (CUDA)}
The goal of our method is to augment more and diverse user utterances that fit the dialogue context, and then the augmented data can help DST models learn better.
 More formally, given a system utterance $U_t^{\mathrm{sys}}$ in the turn $t$ and dialogue history $H_{t-1}$ before this turn, our approach focuses on augmenting a user dialogue act and state, $\hat{A_t}$, and generating the corresponding user utterance $\hat{U}_t^{\mathrm{usr}}$.
Note that each user utterance can be augmented.

\begin{figure*}
    \centering
    \includegraphics[width=\textwidth]{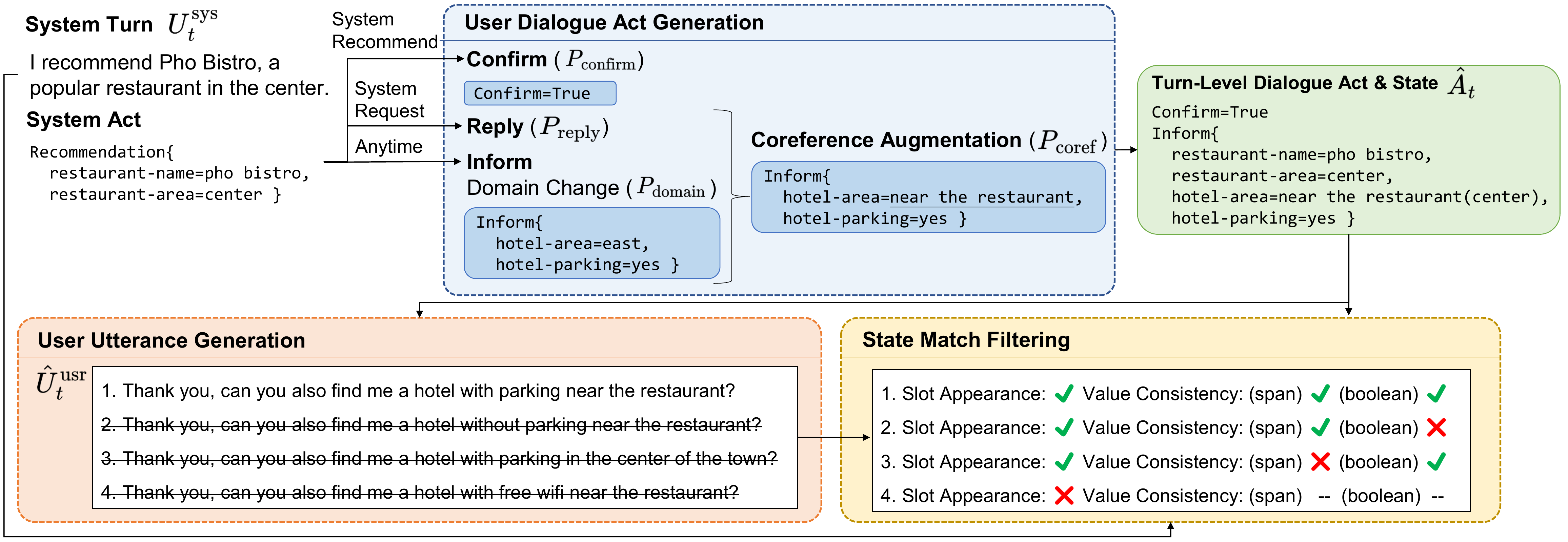}
    \caption{The overview of the proposed CUDA augmentation process.}
    \label{fig:CUDA}
\end{figure*}

To achieve this goal, we propose CUDA with three components illustrated in Figure~\ref{fig:CUDA}:
1) a user dialogue act generation process for producing $\hat{A}_t$,
2) a user utterance generator for producing $\hat{U}_t^{\mathrm{usr}}$,
and
3) a state match filtering process.


\subsection{User Dialogue Act Generation}
Considering that a user dialogue act represents the core meaning of the user's behavior~\cite{goo2018abstractive,yu2021midas}, we focus on simulating reasonable user dialogue acts given the system context for data augmentation.
After analyzing task-oriented user utterances, user behaviors contain the following user dialogue acts:
\begin{compactenum}
\item {\bf Confirm}: The system provides recommendation to the user, and the user confirms if accepting the recommended item.
\item {\bf Reply}: The system asks for a user-desired value of the slots, and the user replies the corresponding value.
\item {\bf Inform}: The user directly informs the desired slot values to the system.
\end{compactenum}
\citet{heck-etal-2020-trippy} designed their dialogue state tracker that tackle utterances with different dialogue acts in different ways and achieved good performance, implying that different dialogue acts contain diverse behaviors in the interactions.
To augment more diverse user utterances, we introduce a random process for each user dialogue act.
Unlike the prior work CoCo that did not generate utterance whose dialogue act different from the original one, our design is capable of simulating diverse behaviors for better augmentation illustrated in Figure~\ref{fig:CUDA}.

\paragraph{Confirm} 

When the system provides recommendations, our augmented user behavior has a probability of $P_\text{confirm}$ to accept the recommended values.
When the user confirms the recommendation, the suggested slot values are added to the augmented user dialogue state $\hat{A}_t$ as shown in Figure~\ref{fig:example}.
In the example, the augmented user dialogue act is to confirm the suggested restaurant, and then includes it in the state (\textsf{restaurant-name=pho bistro, restaurant-area=center}).

\paragraph{Reply}
When the system requests a constraint for a specific slot, e.g. ``{\it which area do you prefer?}'', the user has a probability of $P_\text{reply}$ to give the value of the requested slot.
$P_\text{reply}$ may not be 1, because users sometimes revise their previous requests without providing the asked information.

\paragraph{Inform}
In anytime of the conversation, the user can provide the desired slot values to convey his/her preference. 
As shown in the original user utterance of Figure~\ref{fig:example}, the user rejects the recommendation and then directly informs the additional constraints (\textsf{food} and \textsf{time}).
The number of additional informed values is randomly chosen, and then the slots and values are randomly sampled from the pre-defined ontology and dictionary.
Note that the confirmed and replied information cannot be changed during additional informing.
Considering that a user may change the domain within the dialogue,  our algorithm allows the user to change the domain with a probability of $P_\text{domain}$, and then the informed slots and values need to be sampled from the new domain's dictionary. 
The new domain is selected randomly from all the other domains.

\vspace{-1mm}
\paragraph{Coreference Augmentation}

In the generated user dialogue act and state, all informed slot values are from the pre-defined dictionary.
However, it is natural for a user to refer the previously mentioned information, e.g., ``\textit{I am looking for a taxi that can arrive by the time of my reservation}''.
To further enhance the capability of handling coreference, our algorithm has a probability of $P_\text{coref}$ to switch the slot value from the generated user dialogue state.
Since not all slots can be referred, we define a coreference list containing all referable slots and the corresponding referring phrases, e.g., ``\textit{the same area as}'' listed in Appendix~\ref{sec:detail}.
\vspace{2mm}
With the generated user dialogue acts and the system action, we form the corresponding turn-level dialogue act and state based on the confirmed suggestions and referred slot values as shown in the green block of Figure~\ref{fig:CUDA}.

\subsection{User Utterence Generation}
To generate the user utterance associated with the augmented user dialogue act and state, we adopt a pre-trained T5~\cite{JMLR:v21:20-074} and fine-tune it on the MultiWOZ dataset by a language modeling objective formulated below:
\begin{eqnarray}
    \mathcal{L}_\text{gen} = - \sum^{n_t}_{k=1} \mathrm{log} \, p_{\theta}(U^{\mathrm{usr}}_{t,k} \mid U^{\mathrm{usr}}_{t,<k}, U^{\mathrm{sys}}_{t}, H_{t-1}, A_t),\nonumber
\end{eqnarray}
where $U^{\mathrm{usr}}_{t,k}$ denotes the $k$-th token in the user utterance, $H_{t-1}$ represents the all dialogue history before turn $t$, and $A_t$ is the user dialogue act and state in the $t$-th turn.
With the trained generator, we can generate the augmented user utterance by inputting the augmented user dialogue act and state $\hat{A}_t$ as shown in the green block of Figure~\ref{fig:CUDA}.
In decoding, we apply beam search so that we can augment diverse utterances for improving DST.

\subsection{State Match Filtering}

To make sure the generated user utterance well reflects its dialogue state, we propose two modules to check the state matching: a \emph{slot appearance} classifier and a \emph{value consistency} filter, where the former checks if the given slots are included and the latter focuses on ensuring the value consistency between dialogue states and user utterances.

\paragraph{Slot Appearance}

Following \citeauthor{li2020coco}, we employ a BERT-based multi-label classification model to predict whether a slot appears in the given $t$-th turn.
The augmented user utterances are eliminated if they do not contain all slots in the user dialogue state predicted by the model.

\paragraph{Value Consistency}

The slot values can be categorized into: 1) span-based, 2) boolean, and 3) {\it dontcare} values. 
    It is naive to check if the span-based values are mentioned in the utterances, but boolean and {\it dontcare} values cannot be easily identified.
    To handle the slots with boolean and {\it dontcare} values, we propose two slot-gate classifiers motivated by \citet{heck-etal-2020-trippy}.
    Each boolean slot, e.g. {\it internet} or {\it parking}, is assigned to one of the classes in $C_\text{bool} = \{\textit{none}, \textit{dontcare}, \textit{yes}, \textit{no} \}$, while other slots are assigned to one of the classes in $C_\text{span} = \{\textit{none}, \textit{dontcare}, \textit{value}\}$, where {\it value} indicates the span-based value.
    Then for all slots classified as span-based value, we check if all associated values are mentioned in the generated utterance. In addition, we use the coreference keywords, e.g., \textit{same area}, to handle the coreference cases.
We apply BERT~\cite{devlin-etal-2019-bert} to encode the $t$-th turn in a dialogue as:
\begin{equation}
    \begin{aligned}
        R_t^\mathrm{CLS} = \mathrm{BERT}([\mathrm{CLS}] \oplus & U^{\mathrm{sys}}_t \oplus [\mathrm{SEP}] \oplus \\
        & U^{\mathrm{usr}}_t \oplus [\mathrm{SEP}]),
    \end{aligned}\nonumber
\end{equation}
where ${R_t^\mathrm{CLS}}$ denotes the output of the [CLS] token, which can be considered as the summation of the turn $t$.
We then obtain the probability of the value types as
\begin{equation}
    p^{\mathrm{bool}}_{s, t} =\mathrm{softmax}(W^{\mathrm{bool}}_{s} \cdot R_t^\mathrm{CLS} + b^{\mathrm{bool}}_{s}) \in \mathbb{R}^4
    ,\nonumber
\end{equation}
for each boolean slots, and
\begin{equation}
    p^{\mathrm{span}}_{s, t} =\mathrm{softmax}(W^{\mathrm{span}}_{s} \cdot R_t^\mathrm{CLS} + b^{\mathrm{span}}_{s}) \in \mathbb{R}^3
    ,\nonumber
\end{equation}
for each span-based slots. 
Our multi-task BERT-based slot-gate classifier is trained with the cross entropy loss.
The neural-based filters are trained on the original MultiWOZ data, and the prediction performance in terms of slots (for both appearance and value consistency) is 92.9\% in F1 evaluated on the development set.
In our CUDA framework, we apply the trained filters to ensure the quality of the augmented user utterances as shown in Figure~\ref{fig:CUDA}.

\begin{table}[t!]
\centering \small
\begin{tabular}{lrr}
\toprule
\bf Dataset & \bf CUDA & \bf MultiWOZ  \\
\midrule
Span  & 100.00 & 64.61  \\
Confirm (True)  & 5.27 &  5.84   \\
Confirm (False)  &  0.44 & 0.32   \\
Dontcare  & 0.67 & 2.46   \\
Coreference & 8.15 & 3.70 \\
Multi-domain  & 13.10 & 24.48   \\
\midrule
\#Turns & \it 54,855 & \it 69,673  \\
\bottomrule
\end{tabular}
\caption{Slot distribution in user utterances (\%). }
\label{tab:analysis} 
\end{table}

\begin{table}[t!]
\centering \small
\begin{tabular}{lccc}
\toprule
\bf MultiWOZ & \bf TripPy & \bf TRADE & \bf SimpleTOD \\
\midrule
Original & 57.72 & 44.08 & 49.19 \\
VS & 59.48 & 43.76 & \bf{50.50} \\
CoCo & 60.46 & 43.53 & 50.25 \\
CUDA & 61.28$^\dag$ & \bf{44.86}$^\dag$ & 50.14\\
CUDA (-\textit{coref}) & \bf{62.93}$^\dag$ & 42.98 & 49.64  \\
\bottomrule
\end{tabular}
\caption{Joint goal accuracy on MultiWOZ 2.1 (\%). $\dag$ indicates the significant improvement over all baselines with $p < 0.05$.}
\label{tab:multiwoz} 
\end{table}

\section{Experiments}

To evaluate if our augmented data is beneficial for improving DST models, we perform three popular trackers, TRADE~\cite{wu-etal-2019-transferable}, SimpleTOD~\cite{hosseini2020simple}, and TripPy~\cite{heck-etal-2020-trippy}, on MultiWOZ 2.1~\cite{eric-etal-2020-multiwoz}.


\subsection{Experimental Setting}

Our CUDA generator is trained on the training set of MultiWOZ 2.3~\cite{han2020multiwoz} due to its additional \textit{coreference} labels. Note that all dialogues are the same as MultiWOZ 2.1.
We then generate the augmented dataset for the training set of MultiWOZ 2.1 for fair comparison with the prior work.
The predifined slot-value dictionary is taken from CoCo's \textit{out-of-domain} dictionary and the defined coreference list is shown in Appendix~\ref{sec:detail}.

In user dialogue act generation, the parameters are set as
$(P_\text{confirm}, P_\text{reply}, P_\text{domain}, P_\text{coref}) = (0.7, 0.9, 0.8, 0.6)$,
which can be flexibly adjusted to simulate different user behaviors.
We report the distribution of slot types in our augmented data and the original MultiWOZ data in Table~\ref{tab:analysis}, where it can be found that our augmented slots cover diverse slot types and the distribution is reasonably similar to the original MultiWOZ.
Different from the prior work, CoCo, which only tackled the span-based slots, our augmented data may better reflect the natural conversational interactions.
Additionally, we perform CUDA with ${P_\text{coref} = 0}$ to check the impact of coreference augmentation.

We train three DST models on the augmented data and evaluate the results using joint goal accuracy.
The compared augmentation baselines include value substitution (VS) and CoCo~\cite{li2020coco} with the same setting. 

\subsection{Effectiveness of CUDA-Augmented Data}


Table \ref{tab:multiwoz} shows that CUDA significantly improves TripPy and TRADE results by 3.6\% and 0.8\% respectively on MultiWOZ, and even outperforms the prior work CoCo.
In addition, our CUDA augmentation process has 78\% success rate, while CoCo only has 57\%, demonstrating the efficiency of our augmentation method and the great data utility.
Interestingly, CUDA without \textit{coreference} achieves slightly better performance for TripPy while the performance of TRADE and SimpleTOD degrade, achieving the new state-of-the-art performance on MultiWOZ 2.1.
The probable reason is that TripPy already handles coreference very well via its refer classification module, so augmenting coreference cases may not help it a lot.
In contrast, other generative models (TRADE and SimpleTOD) can benefit more from our augmented coreference cases.
Another reason may be the small distribution of coreference slots in MultiWOZ shown as Table~\ref{tab:analysis}, implying that augmented data with too many coreference slots does not align well with the original distribution and hurts the performance.

\begin{table}[t!]
\centering \small
\begin{tabular}{lccc}
\toprule
\bf CoCo+(rare) & \bf TripPy & \bf TRADE & \bf SimpleTOD \\
\midrule
Original & 28.38 & 16.65 & 19.20\\
VS & 39.42 & 16.42 & 26.26 \\
CUDA & \bf 48.83 & \bf 17.79 & \bf 29.32  \\
CUDA (-\textit{coref}) & 48.67 & 16.80 & 28.66  \\
\midrule
CoCo & \it 56.50 & \it 18.01 & \it 30.60  \\
\bottomrule
\end{tabular}
\caption{Joint goal accuracy on CoCo+ (rare) (\%). }
\label{tab:coco+} 
\end{table}

\subsection{Robustness to Rare Cases}

We also evaluate our models on \textit{CoCo+ (rare)}\footnote{CoCo+ (rare) applies \textit{CoCo} and \textit{value substitution (VS)} with a \textit{rare} slot-combination dictionary.}, a test set generated by CoCo's algorithm~\cite{li2020coco}, to examine model robustness under rare scenarios.
Table~\ref{tab:coco+} presents the results on CoCo+ (rare), which focuses rare cases for validating the model's robustness.
It is clear that the model trained on our augmented data shows better generalization compared with the one trained on the original MultiWOZ data, demonstrating the effectiveness on improving  robustness of DST models.
The performance of CoCo is listed as reference, because comparing with its self-generated data is unfair.

\subsection{Slot Performance Analysis}

\begin{figure}[t]
    \centering
    \includegraphics[width=\linewidth]{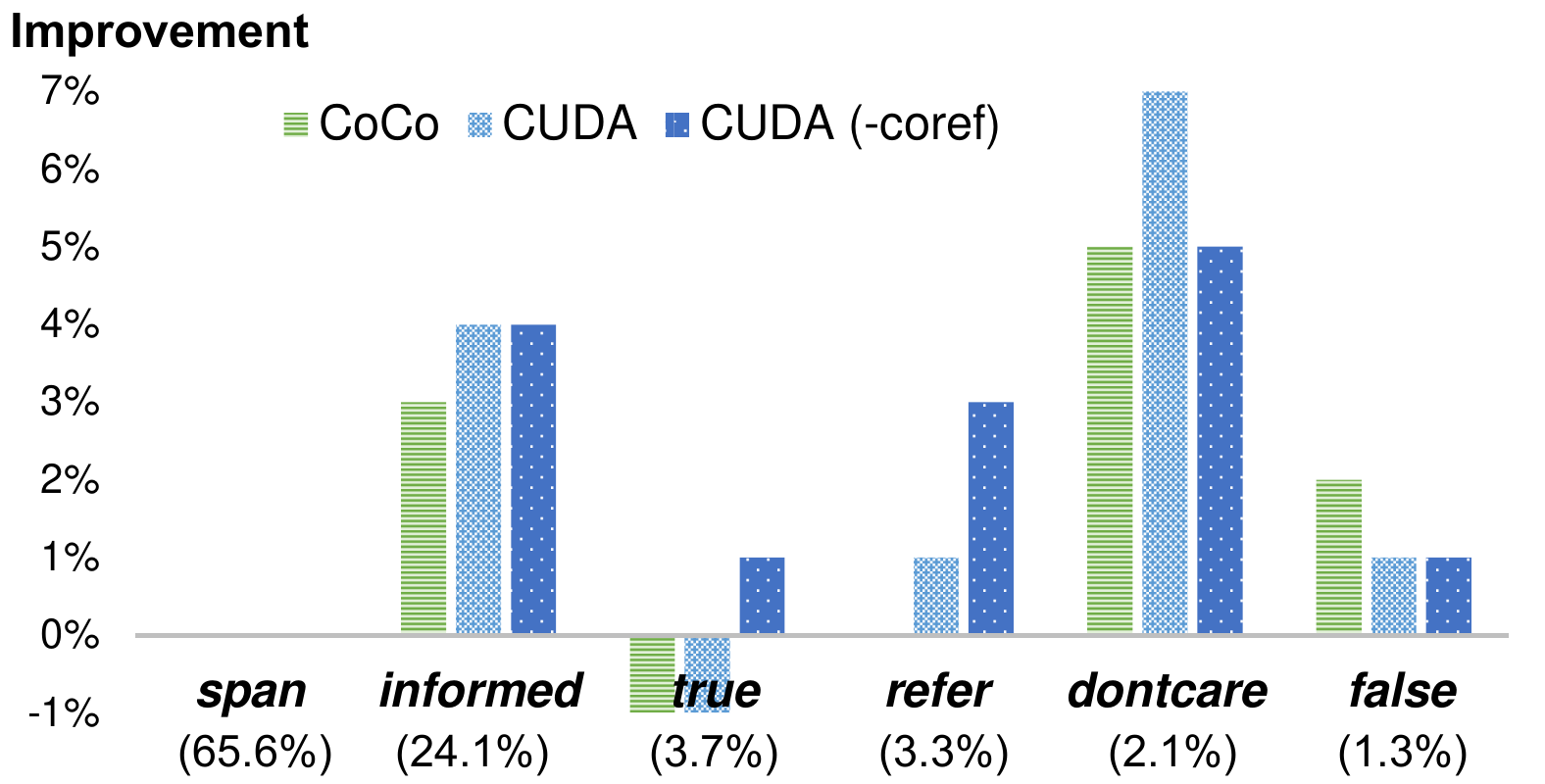}
    \vspace{-3mm}
    \caption{Performance gain across slots on TripPy.}
    \label{fig:slot-analysis}
    \vspace{-1mm}
\end{figure}

To further investigate the efficacy for each slot type, Figure~\ref{fig:slot-analysis} presents its performance gain on TripPy.
Comparing with CoCo, CUDA improves more on \textit{informed}, \textit{refer}, and \textit{dontcare} slots.
It implies that CUDA augments diverse user dialogue acts for helping \textit{informed} and \textit{refer}, and the proposed slot-gate can better ensure value consistency for improving \textit{dontcare} slots, even though they are rare cases in MultiWOZ.
Our model can also keep the same performance for frequent \textit{span} slots, demonstrating great generalization capability across diverse slot types from our controllable augmentation.
The qualitative study can be found in Appendix~\ref{sec:qual}.

\section{Conclusion}

We introduce a generalized data augmentation method for DST by utterance generation with controllable user dialogue act augmentation.
Experiments show that our approach improves results of multiple state trackers and achieves state-of-the-art performance on MultiWOZ 2.1.
Further study demonstrates that trackers' robustness and generalization capabilities can be improved by diverse generation covering different user behaviors.

\section*{Acknowledgements}
We thank reviewers for their insightful comments.
This work was financially supported from the Young Scholar Fellowship Program by Ministry of Science and Technology (MOST) in Taiwan, under Grants 111-2628-E-002-016 and 111-2634-F-002-014.

\bibliography{anthology,custom}
\bibliographystyle{acl_natbib}


\appendix

\section{Reproducibility}
\label{sec:detail}
Our CUDA generator is trained on the training set of MultiWOZ 2.3~\cite{han2020multiwoz} due to its additional \textit{coreference} labels. Note that all dialogues are the same as MultiWOZ 2.1.
We then generate the augmented dataset using CUDA for the training set of MultiWOZ 2.1 for fair comparison with the prior work.
The predifined slot-value dictionary is taken from CoCo's \textit{out-of-domain} dictionary shown in Table~\ref{tab:slot-value} and the defined coreference list is shown in Table~\ref{tab:coreference-list}.

\section{Qualitative Study}
\label{sec:qual}
\begin{figure*}[h]
    \centering
    \includegraphics[width=\textwidth]{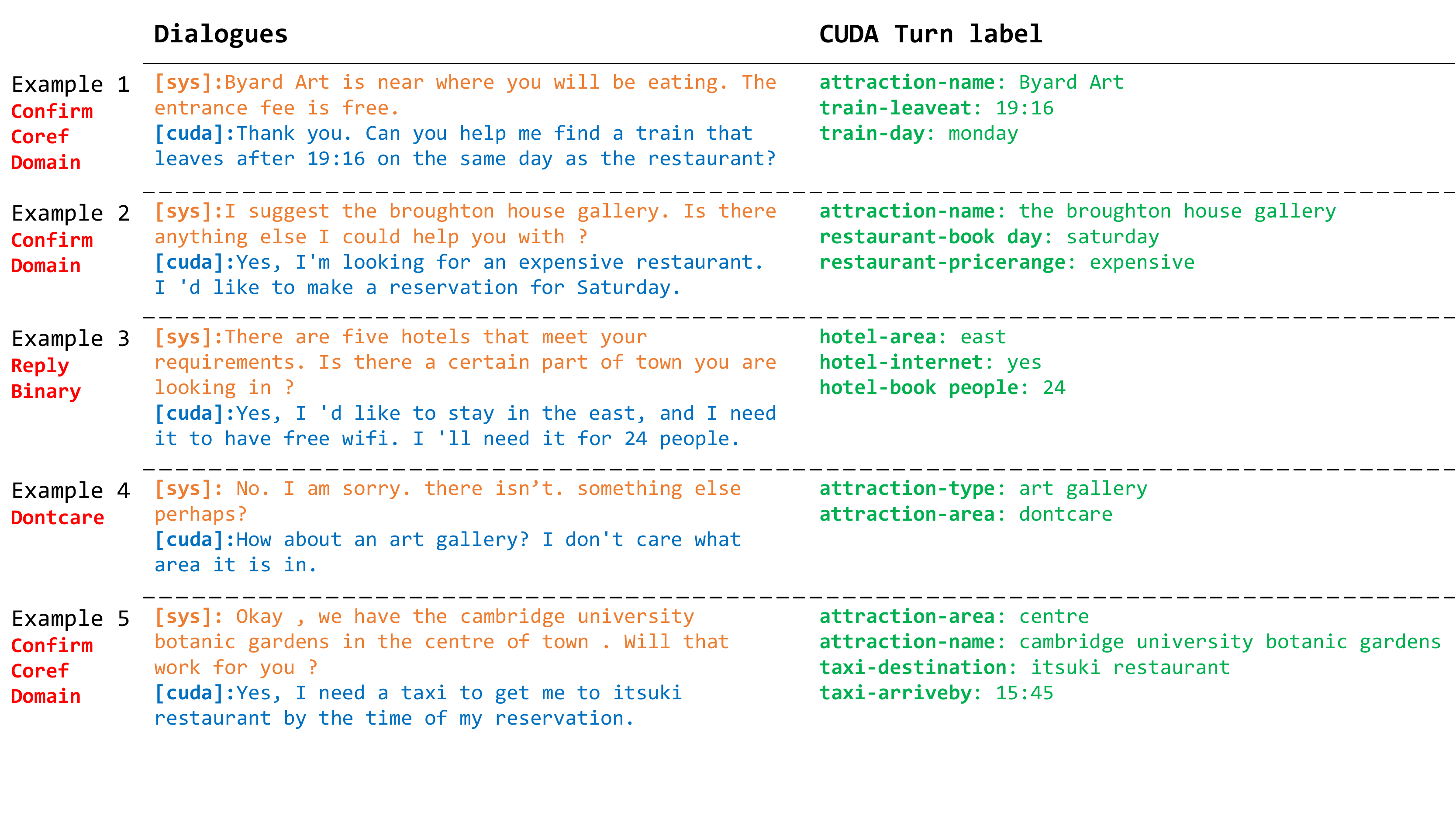}
    \caption{The CUDA-generated examples. The red tags indicate the strategies implemented by CUDA.}
    \label{fig:sample}
\end{figure*}

The augmented data samples are shown in Figure~\ref{fig:sample}.
It can be found that the augmented user utterances can fluently switch the domain and include associated slot values that are aligned well with the dialogue states.

\begin{table*}[]
    \centering
    \small
    \begin{tabular}{|l|p{13cm}|}
        \hline
         \textbf{Slot Name} & \textbf{Possible Values} \\
        \hline \hline
         \textit{hotel-internet$^\dag$} & [`yes', `no', `dontcare'] \\
        \hline 
         \textit{hotel-type} & [`hotel', `guesthouse'] \\
        \hline 
         \textit{hotel-parking$^\dag$} & [`yes', `no', `dontcare'] \\
        \hline 
         \textit{hotel-price} & [`moderate', `cheap', `expensive'] \\
        \hline 
         \textit{hotel-day} & [`march 11th', `march 12th', `march 13th', `march 14th', `march 15th', `march 16th', `march 17th', `march 18th', `march 19th', `march 20th'] \\
        \hline 
         \textit{hotel-people} & [`20', `21', `22', `23', `24', `25', `26', `27', `28', `29'] \\
        \hline 
         \textit{hotel-stay} & [`20', `21', `22', `23', `24', `25', `26', `27', `28', `29'] \\
        \hline 
         \textit{hotel-area} & [`south', `north', `west', `east', `centre', `dontcare'] \\
        \hline 
         \textit{hotel-stars} & [`0', `1', `2', `3', `4', `5', `dontcare'] \\
        \hline 
         \textit{hotel-name} & [`moody moon', `four seasons hotel', `knights inn', `travelodge', `jack summer inn', `paradise point resort'] \\
        \hline 
         \textit{restaurant-area} & [`south', `north', `west', `east', `centre', `dontcare'] \\
        \hline 
         \textit{restaurant-food} & [`asian fusion', `burger', `pasta', `ramen', `taiwanese', `dontcare'] \\
        \hline 
         \textit{restaurant-price} & [`moderate', `cheap', `expensive', `dontcare'] \\
        \hline 
         \textit{restaurant-name} & [`buddha bowls', `pizza my heart', `pho bistro', `sushiya express', `rockfire grill', `itsuki restaurant'] \\
        \hline 
         \textit{restaurant-day} & [`monday', `tuesday', `wednesday', `thursday', `friday', `saturday', `sunday'] \\
        \hline 
         \textit{restaurant-people} & [`20', `21', `22', `23', `24', `25', `26', `27', `28', `29'] \\
        \hline 
         \textit{restaurant-time} & [`19:01', `18:06', `17:11', `19:16', `18:21', `17:26', `19:31', `18:36', `17:41', `19:46', `18:51', `17:56',  `7:00 pm', `6:07 pm', `5:12 pm', `7:17 pm', `6:17 pm', `5:27 pm', `7:32 pm', `6:37 pm', `5:42 pm', `7:47 pm', `6:52 pm', `5:57 pm', `11:00 am', `11:05 am', `11:10 am', `11:15 am', `11:20 am', `11:25 am', `11:30 am', `11:35 am', `11:40 am', `11:45 am', `11:50 am', `11:55 am'] \\
        \hline 
         \textit{restaurant-food} & [`asian fusion', `burger', `pasta', `ramen', `taiwanese', `dontcare'] \\
        \hline 
         \textit{taxi-arrive} & [`17:26', `19:31', `18:36', `17:41', `19:46', `18:51', `17:56', `7:00 pm', `6:07 pm', `5:12 pm', `7:17 pm', `6:17 pm', `5:27 pm', `11:30 am', `11:35 am', `11:40 am', `11:45 am', `11:50 am', `11:55 am'] \\
        \hline 
         \textit{taxi-leave} & [`19:01', `18:06', `17:11', `19:16', `18:21', `7:32 pm', `6:37 pm', `5:42 pm', `7:47 pm', `6:52 pm', `5:57 pm', `11:00 am', `11:05 am', `11:10 am', `11:15 am', `11:20 am', `11:25 am'] \\
        \hline 
         \textit{taxi-depart} & [`moody moon', `four seasons hotel', `knights inn', `travelodge', `jack summer inn', `paradise point resort'] \\
        \hline 
         \textit{taxi-dest} & [`buddha bowls', `pizza my heart', `pho bistro', `sushiya express', `rockfire grill', `itsuki restaurant'] \\
        \hline 
         \textit{train-arrive} & [`17:26', `19:31', `18:36', `17:41', `19:46', `18:51', `17:56', `7:00 pm', `6:07 pm', `5:12 pm', `7:17 pm', `6:17 pm', `5:27 pm', `11:30 am', `11:35 am', `11:40 am', `11:45 am', `11:50 am', `11:55 am'] \\
        \hline 
         \textit{train-leave} & [`19:01', `18:06', `17:11', `19:16', `18:21', `7:32 pm', `6:37 pm', `5:42 pm', `7:47 pm', `6:52 pm', `5:57 pm', `11:00 am', `11:05 am', `11:10 am', `11:15 am', `11:20 am', `11:25 am'] \\
        \hline 
         \textit{train-depart} & [`gilroy', `san martin', `morgan hill', `blossom hill', `college park', `santa clara', `lawrence', `sunnyvale'] \\
        \hline 
         \textit{train-dest} & [`mountain view', `san antonio', `palo alto', `menlo park', `hayward park', `san mateo', `broadway', `san bruno'] \\
        \hline 
         \textit{train-day} & [`march 11th', `march 12th', `march 13th', `march 14th', `march 15th', `march 16th', `march 17th', `march 18th', `march 19th', `march 20th'] \\
        \hline 
         \textit{train-people} & [`20', `21', `22', `23', `24', `25', `26', `27', `28', `29'] \\
        \hline 
         \textit{attraction-area} & [`south', `north', `west', `east', `centre', `dontcare'] \\
        \hline 
         \textit{attraction-name} & [`grand canyon', `golden gate bridge', `niagara falls', `kennedy space center', `pike place market', `las vegas strip'] \\
        \hline 
         \textit{attraction-type} & [`historical landmark', `aquaria', `beach', `castle', `art gallery', `dontcare'] \\
        \hline 
    \end{tabular}
    \caption{The pre-defined slot-value dictionary, where $\dag$  indicates a binary slot.}
    \label{tab:slot-value}
\end{table*}

\begin{table*}[]
    \centering
    \small
    \begin{tabular}{|l|l|p{9cm}|}
        \hline
         \textbf{Slot Name} & \textbf{Referred Slot Name} & \textbf{Referred Key Value} \\
        \hline \hline
         \multirow{1}{*}{\textit{hotel-price}} & \textit{restaurant-price} & [`same', `same price', `same price range'] \\
        \hline 
         \multirow{2}{*}{\textit{hotel-day}} & \textit{train-day} & [`same', `same day'] \\
         \cline{2-3}
         & \textit{restaurant-day} & [`same', `same day'] \\
        \hline  
         \multirow{2}{*}{\textit{hotel-people}} & \textit{train-people} & [`same', `same group', `same party'] \\
         \cline{2-3}
         & \textit{restaurant-people} & [`same', `same group', `same party'] \\
        \hline 
         \multirow{2}{*}{\textit{hotel-area}} & \textit{restaurant-area} & [`same', `same area', `same part', `near the restaurant'] \\
         \cline{2-3}
         & \textit{attraction-area} & [`same', `same area', `same part', `near the attraction'] \\
        \hline 
         \multirow{2}{*}{\textit{restaurant-area}} & \textit{hotel-area} & [`same', `same area', `same part', `near the hotel'] \\
         \cline{2-3}
         & \textit{attraction-area} & [`same', `same area', `same part', `near the attraction'] \\
        \hline 
         \multirow{1}{*}{\textit{restaurant-price}} & \textit{hotel-price} & [`same', `same price', `same price range'] \\
        \hline 
         \multirow{2}{*}{\textit{restaurant-day}} & \textit{train-day} & [`same', `same day'] \\
         \cline{2-3}
         & \textit{hotel-day} & [`same', `same day'] \\
        \hline 
         \multirow{2}{*}{\textit{restaurant-people}} & \textit{train-people} & [`same', `same group', `same party'] \\
         \cline{2-3}
         & \textit{hotel-people} & [`same', `same group', `same party'] \\
        \hline 
         \multirow{3}{*}{\textit{taxi-depart}} & \textit{hotel-name} & [`the hotel'] \\
         \cline{2-3}
         & \textit{restaurant-name} & [`the restaurant'] \\
         \cline{2-3}
         & \textit{attraction-name} & [`the attraction'] \\
        \hline 
         \multirow{3}{*}{\textit{taxi-dest}} & \textit{hotel-name} & [`the hotel'] \\
         \cline{2-3}
         & \textit{restaurant-name} & [`the restaurant'] \\
         \cline{2-3}
         & \textit{attraction-name} & [`the attraction'] \\
        \hline 
         \multirow{1}{*}{\textit{taxi-arrive}} & \textit{restaurant-time} & [`the time of my reservation', `the time of my booking'] \\
        \hline 
         \multirow{2}{*}{\textit{train-day}} & \textit{restaurant-day} & [`same', `same day'] \\
         \cline{2-3}
         & \textit{hotel-day} & [`same', `same day'] \\
        \hline 
         \multirow{2}{*}{\textit{train-people}} & \textit{restaurant-people} & [`same', `same group', `same party'] \\
         \cline{2-3}
         & \textit{hotel-people} & [`same', `same group', `same party'] \\
        \hline 
         \multirow{2}{*}{\textit{attraction-area}} & \textit{hotel-area} & [`same', `same area', `same part', `near the hotel'] \\
         \cline{2-3}
         & \textit{restaurant-area} & [`same', `same area', `same part', `near the restaurant'] \\
        \hline 
    \end{tabular}
    \caption{The coreference list. The slots that is not referable will not be displayed in the above table.}
    \label{tab:coreference-list}
\end{table*}

\end{document}